\documentclass[runningheads]{llncs}

\usepackage{times}
\usepackage{soul}
\usepackage{url}
\usepackage[hidelinks]{hyperref}
\usepackage[utf8]{inputenc}
\usepackage[small]{caption}
\usepackage{graphicx}
\usepackage{booktabs}
\usepackage{algorithm}
\usepackage{algorithmic}
\usepackage{bm}
\usepackage{xspace}
\usepackage{xfrac}
\usepackage{diagbox}

\begin{document}

\newif\ifjournal
\journalfalse 

\title{Strategy Proof Mechanisms for \\Facility Location at Limited Locations}

\author{Toby Walsh}

\institute{UNSW Sydney and CSIRO Data61}

\newtheorem{mytheorem}{Theorem}
\newtheorem{myproposition}{Proposition}
\newtheorem{myexample}{Example}

\newcommand{\myproof}{\noindent {\bf Proof:\ \ }}
\newcommand{\mymax}{\mbox{\rm max}}
\newcommand{\mymin}{\mbox{\rm min}}
\newcommand{\myqed}{\mbox{$\diamond$}}
\newcommand{\myOmit}[1]{}

\newcommand{\myblacksquare}{$\blacksquare$}

\newcommand{\innerpoint}{\mbox{\sc InnerPoint}\xspace}
\newcommand{\extendedendpoint}{\mbox{\sc ExtendedEndPoint}\xspace}
\newcommand{\percentile}{\mbox{\sc Percentile}\xspace}
\newcommand{\mymedian}{\mbox{\sc Median}\xspace}
\newcommand{\mymidpoint}{\mbox{\sc MidPoint}\xspace}
\newcommand{\mymidpoints}{\mbox{\sc MidPoint}\xspace}
\newcommand{\mymidornearest}{\mbox{\sc MidOrNearest}\xspace}
\newcommand{\mydmidornearest}{\mbox{$\mbox{\sc MidOrNearest}^*$}\xspace}
\newcommand{\mydmidpoint}{\mbox{$\mbox{\sc MidPoint}^*$}\xspace}
\newcommand{\mydmidpoints}{\mbox{$\mbox{\sc MidPoint}^*$}\xspace}
\newcommand{\myendpoint}{\mbox{\sc EndPoint}\xspace}
\newcommand{\myendpoints}{\mbox{\sc EndPoint}\xspace}
\newcommand{\median}{\mbox{\sc Median}\xspace}
\newcommand{\leftmost}{\mbox{\sc Leftmost}\xspace}
\newcommand{\rightmost}{\mbox{\sc Rightmost}\xspace}
\newcommand{\jleftkright}{\mbox{\sc $j$Left$k$Right}\xspace}
\newcommand{\twopeaks}{\mbox{\sc TwoPeaks}\xspace}
\newcommand{\twoleftpeaks}{\mbox{\sc TwoLeftPeaks}\xspace}
\newcommand{\tworightpeaks}{\mbox{\sc TwoRightPeaks}\xspace}
\newcommand{\threepeaks}{\mbox{\sc ThreePeaks}\xspace}
\newcommand{\threeleftpeaks}{\mbox{\sc ThreeLeftPeaks}\xspace}
\newcommand{\threerightpeaks}{\mbox{\sc ThreeRightPeaks}\xspace}
\newcommand{\firstsecond}{\mbox{\sc CapSD}\xspace}
\newcommand{\firstsecondthird}{\mbox{\sc CapaSD}\xspace}
\newcommand{\dmedian}{\mbox{$\mbox{\sc Median}^*$}\xspace}
\newcommand{\dleftmost}{\mbox{$\mbox{\sc Leftmost}^*$}\xspace}
\newcommand{\drightmost}{\mbox{$\mbox{\sc Rightmost}^*$}\xspace}
\newcommand{\ddleftmost}{\mbox{$\mbox{\sc Leftmost}^-$}\xspace}
\newcommand{\mydendpoint}{\mbox{$\mbox{\sc EndPoint}^*$}\xspace}
\newcommand{\mydendpoints}{\mbox{$\mbox{\sc EndPoints}^*$}\xspace}
\newcommand{\myddendpoints}{\mbox{$\mbox{\sc EndPoints}_*$}\xspace}
\newcommand{\dgmv}{\mbox{$\mbox{\sc GenMedian}^*$}\xspace}
\newcommand{\gmv}{\mbox{$\mbox{\sc GenMedian}$}\xspace}
\newcommand{\twodquarter}{\mbox{$\mbox{\sc QuarterPoints}^*$}\xspace}
\newcommand{\twodthirds}{\mbox{$\mbox{\sc ThirdPoints}^*$}\xspace}
\newcommand{\twoquarter}{\mbox{$\mbox{\sc QuarterPoints}$}\xspace}
\newcommand{\twothirds}{\mbox{$\mbox{\sc ThirdPoints}$}\xspace}
\newcommand{\fixedm}{\mbox{$\mbox{\sc UniformFixed}$}\xspace}
\newcommand{\mypmb}[1]{\mbox{$\bm{#1}$}}

\maketitle

\begin{abstract}
Most studies of facility location problems permit a facility
to be located at any position. 
In practices, this may not be possible.
For instance, we might have to limit facilities
to particular locations
such as at highway exits, close to bus stops, or on 
empty building lots.
We consider here the impact of such constraints on the location of 
facilities on the performance of strategy proof mechanisms for
facility location. We study six different 
objectives: 
the total or maximum distance agents must  travel, 
the utilitarian or egalitarian welfare, 
and the total or minimum satisfaction of agents (satisfaction is a normalized
form of utility). 
We show that limiting the location of a facility
makes all six objectives harder to approximate. 
For example, we prove that the median mechanism modified suitably to locate
the facility only at a feasible location
is strategy proof and 3-approximates both the optimal
total distance and the optimal maximum distance. In fact, this is optimal
as no deterministic and strategy proof mechanism can better
approximate the total or maximum distances. 
This contrasts with the setting where the facility can be located
anywhere, and the median mechanism returns the optimal total distance
and 2-approximates the maximum distance. 
\end{abstract}

\section{Introduction}

The facility location problem has been studied using tools 
from a wide variety of fields such as 
AI
(e.g. \cite{ptacmtec2013,gnpijcai18}).
Operations Research
(e.g. \cite{flp,mnwflp}), and Game Theory
(e.g. \cite{proportional,egktps2011}). 
Our goal here is to design mechanisms that locate
the facility in a way that the agents
have no incentive to mis-report their true locations. 
Facility location models many practical problems including 
locating bus or tram stops, schools, playgrounds, telephone exchanges, 
mobile phone masts, recycling centres, electric car charging
locations, shared cars, 
power plants, electricity substations, doctors, chemists, fire stations, 
and hospitals to serve a community. In many of these real world
settings, facilities
may be limited in where they can be located. For example, 
a warehouse might need to be constrained to be near to the railway,
or an ambulance station close to a highway. 
Our contribution is to demonstrate that such
constraints on the location of a facility
make it harder to design strategy proof
mechanisms which provide high quality solutions. 
We measure the quality of the solution in six different 
ways: total or maximum distance of the agents to the facility,
the utilitarian and egalitarian welfare, and the social or minimum
satisfaction. 

From a technical sense, limiting the location of a facility might 
appear to change little the facility location problem. 
We merely need to limit the space of mechanisms 
to the strict subset of mechanism which only locate facilities at 
feasible locations.
We can therefore immediately inherit many impossibility results. 
For instance, since there is no deterministic and strategy proof mechanism for the 
facility problem which minimizes the maximum distance an agent travels when
the facility can be located anywhere, it follows
quickly that there is no such mechanism when the facility is limited in its
location.
The mechanisms excluded because
they locate facilities at infeasible locations are often precisely
those with good normative properties. 
Our contribution here is to show restricting
mechanisms to locate facilities only at feasible locations
often increases approximation ratios, irrespective of whether
the objective is distance, welfare or satisfaction. 
However, the extent to which approximation ratios increases
depends very much on the objective and the problem.
For example, the lower bound on the best possible approximation
ratio of the optimal egalitarian welfare
increases from $\frac{3}{2}$ to unbounded when we 
limit the feasible location of a facility. 
On the other hand, the best possible approximation
ratio of the optimal utilitarian welfare only triples in
this case. Our results are summarized in Tables 1 and 2. 

\begin{table}[htb]
\begin{center} 
{\scriptsize
\begin{tabular}{|c||c|c|c|c|c|c|} \hline \
\diagbox{mechanism}{measure} & 
\begin{tabular}{c} total  \\ distance \end{tabular} & 
\begin{tabular}{c} max \\ distance \end{tabular}  &
\begin{tabular}{c} utilitarian  \\ welfare \end{tabular} & 
\begin{tabular}{c} egalitarian \\ welfare \end{tabular}  &
\begin{tabular}{c} social \\ satisfaction\end{tabular} & 
\begin{tabular}{c} min \\ satisfaction \end{tabular}  \\ \hline \hline
lower bound & 3 [1] & $\mypmb{3}$ [2] & $\mypmb{3}$ [1] & $\mypmb{\infty}$
                                                      [$\frac{3}{2}$] &
                                                            $\mypmb{\infty}$
                                                            [1.07] & 
$\mypmb{\infty}$ ($\frac{4}{3}$)  \\ \hline
\dmedian & 3 [1] & $\mypmb{3}$ [2] & $\mypmb{3}$ [1] & $\mypmb{\infty}$
                                                      [$\infty$] &
                                                            $\mypmb{\infty}$
$\mypmb{[\frac{3}{2}]}$ & 
$\mypmb{\infty}$ [$\infty$]  \\ 
\hline \hline
\end{tabular}
}
\end{center}
\caption{Approximation ratios achievable by deterministic
and strategy proof mechanisms 
for the single facility
location problem at limited locations. 
{\bf Bold} for results proved here.  [Numbers]
in brackets are the approximation ratios achieved for
when the facility
can be located anywhere.}\label{tab1}
\end{table}

\begin{table}[htb]
\begin{center} 
{\scriptsize
\begin{tabular}{|c||c|c|c|c|c|c|} \hline \
\diagbox{mechanism}{measure} & 
\begin{tabular}{c} total  \\ distance \end{tabular} & 
\begin{tabular}{c} max \\ distance \end{tabular}  &
\begin{tabular}{c} utilitarian  \\ welfare \end{tabular} & 
\begin{tabular}{c} egalitarian \\ welfare \end{tabular}  &
\begin{tabular}{c} social \\ satisfaction\end{tabular} & 
\begin{tabular}{c} min \\ satisfaction \end{tabular}  \\ \hline
 & & & & & & \\
\mydendpoint & 2n-3 [n-2] & 3 [2] & $\mypmb{2}$ [2] & $\mypmb{\frac{3}{2}}$
                                                      [$\frac{3}{2}$] &
                                                            $\mypmb{\frac{3n}{4}-\frac{1}{2}}$
[$\frac{n}{2}-\frac{1}{4}$] & 
$\mypmb{\infty}$ [$\infty$]  \\ 
& & & & & & \\
\hline 
\end{tabular}
}
\end{center}

\caption{Summary of approximation ratios achieved by \mydendpoint\
mechanism for the two facility
location problem at limited locations. 
{\bf Bold} for results proved here.
(Numbers) in brackets are the approximation ratios achieved by
the corresponding \myendpoint\ mechanism when the two facilities
can be located anywhere.}\label{tab2}
\end{table}

\section{Related work}

We follow the line of work initiated by 
Procaccia and Tennenholtz \cite{ptacmtec2013}
that looks to resolve the inherent tension in designing 
mechanisms that are strategy proof and effective
by identifying strategy proof mechanisms that are guaranteed
to return solutions within some constant factor
of optimal. 
The most related prior work to ours is by Feldman, Fiat, Golomb
\cite{ffgec2016}. This also considers facility location
problems where the facility is restricted
to limited locations. There is, however, a critical difference with
this work. This earlier work restricted analysis to 
a single objective (sum of distances), while here we consider
six objectives (sum/maximum distance, utilitarian/egalitarian
welfare, social/minimum satisfaction). Our results 
show that approximation ratios that can be achieved
depend critically on the objective chosen. For instance, when
the facility is restricted to limited locations, 
deterministic and strategy proof mechanisms can 3-approximate
the optimal utilitarian welfare. However, no deterministic
and strategy proof mechanism has a bounded
approximation ratio for the egalitarian welfare. 
By contrast, if we consider a different but related objective to the
egalitarian welfare such 
as the maximum distance an agent travels, then deterministic
and strategy proof mechanisms exist which bound the approximation
ratio even when the facility is restricted to limited locations. 
In addition, when the facility is unrestricted, there exists
a deterministic and strategy proof mechanism that can $\frac{3}{2}$-approximate
the optimal egalitarian welfare. The choice of objective then reveals
different aspects of the approximability of these facility location
problems. 

One month after this work here first appeared as a preprint, 
Tang, Wang, Zhang and Zhao published a preprint looking independently
at a  special case of this problem in which facilities are limited
to a finite set of locations \cite{flplimit2}.  There are two
significant technical differences between the two studies. First, the work 
of Tang {\it et al.} does not capture
the more general setting here where the facility is limited to a
set of subintervals. In their work, a facility is limited to a finite
set of locations. Their model cannot then describe a setting where,
for example, a school must be within 500m of one of the
neighbourhood bus stops as the feasible set 
is not finite. 
In addition, the work of Tang {\it et al.}, like the work of Feldman
{\it et al.}, only considers two objectives:
total and maximum cost. 
Here we consider four additional objectives: utilitarian and egalitarian
welfare, as well as social and minimum satisfaction. Our results show that
we can achieve very different approximation ratios with these
different objectives. Indeed, for many of these new objectives, we 
cannot achieve a bounded approximation ratio with deterministic and
strategy proof mechanisms. 

As in much previous work on mechanism design for facility location
(e.g. \cite{ptacmtec2013}), we consider the one-dimensional setting.
This models a number of real world problems
such as locating shopping centres along a highway, or
ferry stops along a river. There are
also various non-geographical settings that can be viewed
as one-dimensional facility location problems (e.g. choosing
the temperature of a classroom, or the tax rate for property
transactions). 
In addition,
we can use mechanisms for the one-dimensional
facility location problem in more complex settings (e.g. we can decompose the 
2-d rectilinear problem into a pair of 1-d problems).
Finally, results about mechanisms for the one-dimensional problem
can inform the results about mechanisms for more complex metrics. 
For instance, lower bounds on the performance
of strategy proof mechanisms for the 1-d problem provide lower bounds for the 2-d problem. 

\section{Formal background}

We have $n$ agents located on $[0,1]$, and
wish to locate one or more facilities also on $[0,1]$ to serve all the
agents. Agent $i$ is at location $x_i$. Without loss of generality, 
we suppose agents are ordered so that $x_1 \leq \ldots \leq x_n$. 
A solution is a location $y_j$ for each facility $j$. Agents
are served by their nearest facility. 
We consider six different performance measures: total or maximum
distance, utilitarian or egalitarian welfare,
and social or minimum satisfaction.

The total distance is $\sum_{i=1}^n \mymin_j | x_i - y_j|$. The 
maximum distance is $\mymax_{i=1}^n \mymin_j | x_i - y_j|$ . 
We suppose the utility $u_i$ of agent $i$ is inversely related
to its distance from the facility serving it. More precisely, $u_i = 1
- \mymin_j |x_i - y_j|$. 
Utilities are, by definition, in $[0,1]$. 
The utilitarian welfare is the sum of the utilities of the
individual agents, $\sum_{i=1}^n u_i$. The egalitarian 
welfare is the minimum utility of any agent, $\mymin_{i=1}^n u_i$. 

In \cite{flprevisit,flpdesire}, normalized utilities called ``happiness
factors'' are introduced. The happiness $h_i$ of agent $i$ is 
$h_i = 1 - \mymin_j \frac{|x_i - y_j}{d_{max}^i}$ where $d_{max}^i$ is the maximum
possible distance agent $i$ may need to travel.
Here $d_{max}^i =\mymax(x_i,1-x_i)$. 
Note that the happiness of an agent is, by definition, 
in $[0,1]$. 
The social satisfaction is then the sum of the happinesses of the
individual agents, $\sum_{i=1}^n h_i$. The minimum
satisfaction is the minimum happiness of any agent, $\mymin_{i=1}^n h_i$. 
Our goal is to optimize one of the distance, 
welfare or satisfaction objectives. 

We consider some particular mechanisms for facility location.
Many are based on the function $median(z_1, \ldots, z_p)$ 
which returns $z_i$ where $|\{ j | z_j < z_i\}| < \lceil \frac{p}{2} \rceil$
and $|\{ j | z_j > z_i\}| \leq \lfloor \frac{p}{2} \rfloor$. 
With $n-1$ parameters
$z_1$ to $z_{n-1}$ representing ``phantom'' agents, 
a \gmv\ mechanism locates the
facility at $median(x_1, \ldots, x_n, z_1, \ldots, z_{n-1})$
As we argue shortly, such mechanisms characterize an important class of
strategy proof mechanisms \cite{moulin1980}. 
The \leftmost\
mechanism has parameters $z_i = 0$ for $i \in [1,n)$ and
locates the facility at the location of the leftmost agent. 
The \rightmost\
mechanism has parameters $z_i = 1$ for $i \in [1,n)$ and
locates the facility at the location of the rightmost agent. 
The \mymedian\
mechanism has parameters $z_i = 0$ for $i \leq \lfloor
\frac{n}{2} \rfloor$ and $1$ otherwise,
and locates the facility at the median agent if $n$ is odd, and
the leftmost
of the two median agents if
$n$ is even. 
The \mymidornearest\
mechanism is an instance of \gmv\ with parameters $z_i = \frac{1}{2}$
for $i \in [1,n)$,  
locating the facility at $\frac{1}{2}$ if $x_1 \leq
\frac{1}{2} \leq x_n$, and otherwise at the nearest $x_i$ to
$\frac{1}{2}$.  
\myOmit{
We also consider a mechanism that locates
the facility at a fixed location. 
The \mymidpoint\ mechanism locates
the facility at position $\sfrac{1}{2}$ regardless of the
agents.
}
The \myendpoint\ mechanism locates
one facility with the \leftmost\ mechanism
and another with the \rightmost\ mechanism.

We extend this 
model of facility location problems with constraints on the location
of the facility. In particular, we suppose the interval $[0,1]$ is
decomposed into a set of feasible and disjoint (open or closed) 
sub-intervals, and
the facility must be located within one of 
these sub-intervals. 
Our goal is to see how restricting the feasible locations 
of the facility in this way impacts on the performance
of strategy proof mechanisms. 
Note that unlike \cite{ffgec2016}, 
agents are not limited in where they
can be located. 
We only limit where the facility (and not agents) can be located. 
In particular, we modify mechanisms to ensure
the facility is located at a feasible location. 
For instance, the $\dleftmost$ mechanism
modifies the $\leftmost$ mechanism to locate
the facility at the nearest feasible location to the leftmost
agent. 
The $\drightmost$ mechanism
modifies the $\rightmost$ mechanism in a similar fashion. 
Note that we cannot have agents 
simply report their nearest feasible location as there might be a choice
of such locations. Indeed, many of our
results that approximation guarantees are not
bounded arise because of the difficulty of choosing between the two nearest and  
feasible locations to some optimal but infeasible facility location
in a strategy proof way. 

We consider three desirable properties of 
mechanisms: 
anonymity, Pareto optimality and strategy proofness.
Anonymity is a fundamental fairness property that requires 
all agents to be treated alike. 
Pareto optimality is one of the most fundamental normative properties
in economics. It demands that we cannot improve the solution so
one agent is better off without other agents being worse off. 
Finally, strategy proofness is a fundamental game theoretic property
that ensures agents have no incentive to act strategically and try to
manipulate the mechanisms by mis-reporting their locations. 

More formally, a mechanism is {\em anonymous} iff permuting the agents does not
change the solution. 
A mechanism is {\em Pareto optimal} iff 
it returns solutions that are always Pareto optimal. 
A solution is {\em Pareto optimal} iff there is no
other solution in which one agent travels a strictly shorter distance,
and all other agents travel no greater
distance. A mechanism is {\em strategy proof} 
iff no agent can mis-report and thereby travel a shorter distance. 
For instance, the \median\ mechanism is anonymous,
Pareto optimal and strategy proof. 
Finally, we will consider strategy proof
mechanisms that may approximate the
optimal distance, welfare or satisfaction.
A mechanism achieves an approximation ratio $\rho$ 
iff the solution it returns
is within a factor of $\rho$ times the optimal. In this case, we say that the mechanism
$\rho$-approximates the optimal.

Procaccia and Tennenholtz initiated the study
of designing approximate and strategy proof mechanisms for locating
facilities on a line \cite{ptacmtec2013}. With just one facility,
they argue that the \median\ mechanism is strategy proof and optimal for
the total distance, while the \leftmost\ mechanism is strategy proof and
2-approximates the optimal maximum 
distance, and no deterministic and strategy proof mechanism
can do better.

\section{Single facility, distance approximations}

For a single facility on the line, Moulin proved a seminal
result that any mechanism that is
anonymous, Pareto optimal and strategy proof is a generalized
median mechanism, \gmv\ \cite{moulin1980}.
This locates the facility at the median location
of the $n$ agents and $n-1$ ``phantom'' agents.
We cannot apply Moulin's result directly to our setting
as a \gmv\ mechanism may select
an infeasible location for the facility. Instead, we consider 
the \dgmv\ mechanism which locates a facility at the nearest
feasible location to that returned by a \gmv\ mechanism. 
If there are two nearest and 
equi-distant feasible locations, then 
the \dgmv\ mechanism uses a fixed tie-breaking rule for each infeasible 
interval (e.g,. always use the leftmost of the two nearest
locations). 
Here, as indeed throughout the paper, we suppose a fixed tie-breaking rule to ensure that the 
modified mechanism retains anonymity and strategy proofness. 
However, none of our results on performance guarantees
depend on the 
choice.

The \dmedian\ mechanism is an instance of \dgmv\
which locates the facility at the median agent if it is a feasible
location, and otherwise at the nearest feasible location to the median
agent. 
Mass{\'{o}} and Moreno de Barreda prove that, 
when locating a single facility at limited locations, 
a mechanism is anonymous, Pareto
efficient and strategy proof iff it is 
a \dgmv\ mechanism with at least one phantom
agent at 0 and one at 1 (corollary 2 in \cite{symsp}).  
It follows that the \dmedian\ mechanism
is anonymous, Pareto efficient and strategy proof. 

\subsection{Total distance}

We first consider the objective of minimizing 
the total distance agents travel to be served.
The \dmedian\ mechanism 
3-approximates the optimal total distance (Lemma 21 in \cite{ffgec2016}).
In fact, this is optimal.
No deterministic strategy proof mechanism
can better approximate the total distance in general
(Lemma 19 in \cite{ffgec2016}).
By comparison, when the facility can be located anywhere,
the \mymedian\ mechanism 
is strategy proof and returns the {\em optimal}
total distance. Limiting the feasible
locations of a facility therefore worsens the
performance of the best possible deterministic and strategy proof mechanism.
In particular, the best possible deterministic and strategy proof
mechanism goes from returning an optimal solution to 
3-approximating the optimal total distance. 

\myOmit{ 
The \mymidpoint\ mechanism has an unbounded approximation 
ratio for the total distance. Consider, for example, all agents at 0. 
The \mydmidpoint\ mechanism can do no better when the facility
is restricted to limited locations. 
Finally, we consider the \mymidornearest\ mechanism
as this provides good approximation ratios of 
a number of objectives such as the maximum distance, the egalitarian
welfare and the minimum satisfaction when the location of the facility
is unrestricted.
It is not hard to show that this mechanism $n-1$-approximates the total
distance. The worst case has $n-1$ agents at 0, and 1 at
$\frac{1}{2}$. When the facility is restricted to limited locations,
it is not hard to show that the corresponding \mydmidornearest\
mechanism
$2n-1$-approximates the total distance. The worst
case has $n-1$ agents at 0, and one at $\frac{1}{2}+\epsilon$ for
$\epsilon$ tending to zero, and the facility located only at 0 or 1.
}

\subsection{Maximum distance}

We consider next the objective of minimizing the maximum
distance any agent travels to be served. 
The \dmedian\ mechanism 
also 3-approximates the optimal maximum distance. 

\begin{mytheorem} \label{dmedianmax}
The \dmedian\ mechanism 
3-approximates the optimal maximum distance
for a facility
location problem with limited locations.
\end{mytheorem}
\myproof
There are three cases. In the first case,
the median agent is at a feasible
location. The most an agent 
needs to travel to the facility at the
median agent is then
at most the distance of the 
rightmost agent from the leftmost.
This is at most twice the optimal.
Hence, this 2-approximates
the maximum distance.
In the second case, 
the median agent is not at a feasible
location, and the facility is located
to the left at the nearest
feasible location to the median agent. There
are two sub-cases. In the first sub-case,
the optimal location for the facility
is at this point or even further to the left. 
But this means the facility is located
between the leftmost and rightmost 
agents. Hence, this is again at worst
a 2-approximation of the maximum distance.
in the second sub-case, 
the optimal location for the facility
is to the right, within some 
feasible interval to the right of the 
median agent. 
The worst such sub-case for the approximation ratio is when the 
only feasible locations
for the facility are at 0 or $\frac{2}{3}$, two agents are at
$\frac{1}{3}-\epsilon$ with $\frac{1}{3}  > \epsilon >0$,
and one agent is at $1$. In this sub-case, the \dmedian\
mechanism locates the facility at 0. This gives a maximum distance
of $1$. However, 
the optimal maximum distance of just $\frac{1}{3}+\epsilon$ requires
the facility to be located at $\frac{2}{3}$. The approximation
ratio is thus $\frac{3}{1+3\epsilon}$ which tends to 3 from below 
as $\epsilon$ approaches zero. 
The solution returned is therefore at worst a 3-approximation of the optimal
maximum distance. 
The third  case locates the facility to the right
of the median agent and is symmetric to 
second case. 
\myqed

We contrast this with the setting where the 
facility can be located anywhere, and the \mymedian\ mechanism 
2-approximates the optimal maximum
distance. 
In fact, 
when there are no constraints on where facilities can be located, the \mymedian\ mechanism is optimal
as no deterministic and strategy proof mechanism
can do better than 2-approximate the optimal maximum distance
(Theorem 3.2 of \cite{ptacmtec2013}). 
Restricting the feasible
locations of a facility therefore worsens the
performance of a median mechanism
from a 2-approximation of the optimal maximum distance
to a 3-approximation. 

Can any strategy proof mechanism do better than 3-approximate
the maximum distance when we limit the feasible locations of the
facility? We show that no deterministic and
strategy proof mechanism has a smaller approximation ratio for the optimal maximum distance. 

\begin{mytheorem}
For a facility
location problem with limited locations, 
no deterministic and strategy proof mechanism 
can do better than 3-approximate the optimal maximum distance.
\end{mytheorem}
\myproof
Suppose the only feasible locations
for the facility are at $\frac{1}{4}$ or $\frac{3}{4}$, and there
are two agents, one at $\frac{1}{2}-\epsilon$ and the other at $\frac{1}{2}+\epsilon$
for $\frac{1}{2} > \epsilon >0$. 
There are two cases. In the first case, 
the mechanism locates the facility at $\frac{1}{4}$. 
Suppose the agent at $\frac{1}{2}+\epsilon$ mis-reports their
location as $1$. Since the mechanism is strategy
proof, the location of the facility cannot change.
Consider this new problem with one agent at 
$\frac{1}{2}-\epsilon$ and the other at $1$.
The optimal maximum distance is now $\frac{1}{4}+\epsilon$ with
the facility at $\frac{3}{4}$. However, the solution returned by the mechanism 
has a maximum distance of $\frac{3}{4}$. The approximation 
ratio is $\frac{3}{1+4\epsilon}$. As $\epsilon$ approaches zero, this
tends to 3 from below. Hence the mechanism at best 3-approximates
the total distance. The second case where the facility
is located at $\frac{3}{4}$ rather than $\frac{1}{4}$ is symmetric. 
\myqed

Hence the \dmedian\ mechanism is optimal.
No deterministic and strategy proof
mechanism can do better than 3-approximate the optimal maximum
distance. 

\myOmit{
We consider next 
the \mymidpoint\ mechanism. This has an unbounded approximation 
ratio for the maximum distance. Consider, for example, all agents at 0. 
The \mydmidpoint\ mechanism can do no better when the facility
is restricted to limited locations. 
Finally, we consider the \mymidornearest\ mechanism.
It is not hard to show that this mechanism $2$-approximates the maximum
distance. The worst case has agents at 0 and
$\frac{1}{2}$. When the facility is restricted to limited locations,
it is not hard to show that the corresponding \mydmidornearest\
mechanism
$3$-approximates the maximum distance. The worst
case has agents at $\frac{1}{2}$ and $1$, and the facility located
only at
$\frac{1}{4}+\epsilon$ or $\frac{3}{4}$ for $\epsilon$ tending to zero.
}

\myOmit{
\section{Small infeasible intervals}

When infeasible intervals for the facility are small,
we can provide stronger approximation guarantees on the optimal total
or maximum distance.
We begin with the total distance.
We will show that the \dmedian\ mechanism
has just an additive error in approximating the total distance. 
We say that a mechanism $+\beta$-approximates
the optimal total distance iff 
the total distance agents travel to the nearest facility in the solution returned
is less than or equal to $d_{total}+ \beta$
where $d_{total}$ is the optimal total distance. 
We prove that when the longest infeasible interval is small, 
the \dmedian\ mechanism provides a
solution which approximates the total distance well with 
only a small additive error. 

\begin{mytheorem}
For a facility
location problem with limited locations and $2k+1$ agents, 
if the longest infeasible interval is $\lambda$,
then the \dmedian\ mechanism 
$+k\lambda$-approximates the optimal total distance. 
\end{mytheorem}
\myproof
If the median agent is at a feasible location then the 
\dmedian\ mechanism locates the facility here, and this
gives a solution with optimal total distance.
We therefore need to consider just the setting in which
the median agent is located within an infeasible interval. 
A worst such case for the additive approximation error is when
the only feasible locations for the facility are at 0 or $\lambda$, $k+1$
agents are at $\frac{\lambda}{2}$,  the other $k$ agents are at
$\lambda$,
and the tie-breaking rule locates the facility
at 0. This gives a total distance $\frac{(3k + 1)\lambda}{2}$. However, the optimal
total distance of just $\frac{(k + 1)\lambda}{2}$ requires the facility to be
located at $\lambda$. This gives an additive error of $+k\lambda$. 
note that we can eliminate the dependency on the
tie-breaking rule by having the $k+1$ agents at $\frac{\lambda}{2}-\epsilon$ 
for some small $\epsilon$ tending to zero. 
\myqed

To give a tighter approximation guarantee on
the maximum distance, we introduce a novel approximation measure that bounds the 
additive or multiplicative error. 
We say that a mechanism provides an $( \times \alpha,+\beta)$-approximation
of the optimal maximum distance iff 
the maximum distance any agent travels to the nearest facility in the solution returned
is less than or equal to $\mymax(d_{max} \times \alpha, d_{max}+ \beta)$
where $d_{max}$ is the optimal maximum distance. 

\begin{mytheorem}
For a facility
location problem with limited locations, 
if the longest infeasible interval is $\lambda$,
then the \dmedian\ mechanism provides a 
$(\times 2,+\lambda)$-approximation of the optimal maximum distance. 
\end{mytheorem}
\myproof
In the proof of theorem \ref{dmedianmax}, 
the \dmedian\ mechanism 2-approximates
the optimal maximum distance in every case
considered except for one sub-case of the second
case (and its reflection symmetry). 
In this sub-case, the median agent is at an infeasible
location, the facility is located
to the left at the nearest
feasible location to the median agent, 
and the optimal location for the facility
is to the right, within some 
feasible interval to the right of the 
median agent. 
A worst setting is when the 
only feasible locations
for the facility are at 0 or $\lambda$, two agents are at $\frac{\lambda}{2}$,
one agent is at $\frac{3\lambda}{2}$, and the tie-breaking
rule locates the facility at 0. This gives a maximum distance
of $\frac{3\lambda}{2}$. However, 
the optimal maximum distance of just $\frac{\lambda}{2}$ requires
the facility to be located at $\lambda$. The maximum distance in solution is then 
$+ \lambda$ from the optimal. 
note that we can again eliminate the dependency on the
tie-breaking rule by having the two agents at $\frac{\lambda}{2}-\epsilon$ 
for $\epsilon$ tending to zero. 
\myqed

Thus, if $\lambda$ is small, even if there are
many infeasible intervals, the \dmedian\ mechanism provides close
to an 2-approximation of the optimal maximum distance. This is similar
to the performance guarantee the mechanism provides when facilities can be located anywhere. 
}

\section{Single facility, welfare approximations}

We switch now to considering how well strategy proof mechanisms approximate
the utilitarian or egalitarian welfare.

\subsection{Utilitarian welfare}

With no limits on the location of the facility, 
it is not hard to see that the \mymedian\ mechanism 
is strategy proof and returns the optimal utilitarian welfare. 
However, in our setting, the \mymedian\ mechanism may select
an infeasible location for the facility. We consider  instead
the \dmedian\ mechanism which locates the facility
at the nearest feasible location to the median agent. 

\begin{mytheorem} 
For a facility location problem with limited locations, 
the \dmedian\ mechanism is strategy proof and
3-approximates the optimal utilitarian welfare.
\end{mytheorem}
\myproof
A worst case is when the only feasible locations
for the facility are at 0 or 1, $k$ agents are at $\frac{1}{2}$,
and $k$ more at 1. Suppose 
the tie-breaking rule of the \dmedian\ mechanism 
locates the facility at 0. This gives an utilitarian welfare of
$\frac{k}{2}$ units of utility. However, the optimal utilitarian welfare of $\frac{3k}{2}$
units of utility locates
the facility at $1$. We can eliminate the dependency on the
tie-breaking rule by having the $k$ agents not at $\frac{1}{2}$ 
but at $\frac{1}{2}-\epsilon$ 
for $\epsilon$ tending to zero. 
\myqed

As with minimizing the total distance, the 
\dmedian\ mechanism is optimal. No deterministic
and strategy proof mechanism has a better approximation guarantee.

\begin{mytheorem} \label{utilitarian}
For a facility
location problem with limited locations, 
any deterministic and strategy proof mechanism at best
3-approximates the optimal utilitarian welfare.
\end{mytheorem}
\myproof
Suppose the only feasible locations
for the facility are at 0 or $1$, and there
are two agents, one at $\frac{1}{2}-\epsilon$ and the other at $\frac{1}{2}+\epsilon$
for $\frac{1}{2} > \epsilon >0$.
There are two cases. In the first case, 
the mechanism locates the facility at 0. 
Suppose the agent at $\frac{1}{2}+\epsilon$ mis-reports their
location as $1$. Since the mechanism is strategy
proof, the location of the facility cannot change.
Consider this new problem with one agent at 
$\frac{1}{2}-\epsilon$ and the other at $1$.
The optimal utilitarian welfare is now $\frac{3}{2}-\epsilon$, but
the solution returned by the mechanism 
has an utilitarian welfare of $\frac{1}{2}+\epsilon$. The approximation 
ratio is then $\frac{3-2\epsilon}{1+2\epsilon}$. As $\epsilon$
approaches zero, this tends to 3 from below. 
Hence the mechanism at best 3-approximates
the optimal utilitarian welfare. The second case where the facility
is located at 1 is symmetric. 
\myqed

\myOmit{
The \mymidpoint\ mechanism 2-approximates
the utilitarian welfare as every agents gets at least
utility $\frac{1}{2}$. With all agents at 0 (or 1), the
2-approximation bound is met. 
The \mydmidpoint\ mechanism, on the other hand,
has no bound on its approximation ratio of the
utilitarian welfare. Consider all agents at 0,
and the facility located only at 0 or $1-\epsilon$ for
$\epsilon$ tending to zero. 
Finally, we consider the \mymidornearest\ mechanism. 
It is not hard to show that this mechanism $2$-approximates the 
utilitarian welfare. The worst case has $n-1$ agents at 0, and one at
$\frac{1}{2}$.  
When the facility is restricted to limited locations,
it is not hard to show that the corresponding \mydmidornearest\
mechanism
$2n-1$-approximates the utilitarian welfare. The worst
case has $n-1$ agents at 0, and one at $\frac{1}{2}$, and the 
facility located only at 0 or $1-\epsilon$ for
$\epsilon$ tending to zero. 
}

\subsection{Egalitarian welfare}

We turn now to the egalitarian welfare. 
It is not hard to see that the 
\mymedian\ mechanism
may not bound the approximation ratio of the 
optimal egalitarian welfare. 
Similarly, when the location of facilities is limited, 
the corresponding \dmedian\ mechanism also may approximate the optimal egalitarian 
welfare poorly. 

\begin{mytheorem} 
For a facility
location problem with limited locations, the \dmedian\ mechanism
does not bound the approximation ratio of the optimal egalitarian
welfare. 
\end{mytheorem}
\myproof
Suppose facilities can be at 0, $\frac{1}{2}$ or 1. 
If there are two agents at 0 and one at 1 then
the \dmedian\ mechanism locates the facility at 0. 
The egalitarian welfare of this solution has zero units of utility. However,
the optimal egalitarian welfare has $\frac{1}{2}$ unit of utility
and is achieved by locating the facility at $\frac{1}{2}$. The
approximation ratio is therefore unbounded. 
\myqed

In fact, no deterministic and strategy proof mechanism
for a facility
location problem with limited locations has
a bounded approximation ratio.
This contrasts with the setting where the 
facility can be located anywhere and 
it is possible to show that no deterministic and strategy proof
mechanism can do better than $\frac{3}{2}$-approximate
the optimal egalitarian welfare, and that the
\mymidornearest\ mechanism actually achieves this ratio. 

\begin{mytheorem}
For a facility
location problem with limited locations, 
no deterministic and strategy proof mechanism has a bounded
approximation ratio of the optimal egalitarian welfare.
\end{mytheorem}
\myproof
Suppose the only feasible locations
for the facility are at 0 or $1$, and there
are two agents, one at $\frac{1}{4}$ and the other at $\frac{3}{4}$.
There are two cases. In the first case, 
the mechanism locates the facility at 0. 
Suppose the agent at $\frac{3}{4}$ mis-reports their
location as $1$. Since the mechanism is strategy
proof, the location of the facility cannot change.
Consider this new problem with one agent at 
$\frac{1}{4}$ and the other at $1$.
The optimal egalitarian welfare is now $\frac{1}{4}$ units of
utility, but the solution returned by the mechanism 
has an egalitarian welfare of zero units of utility. The approximation 
ratio is then unbounded. The second case where the facility
is located at 1 is symmetric. 
\myqed

\section{Single facility, satisfaction approximations}

We consider next how well strategy proof mechanisms approximate
the optimal social or minimum satisfaction. 

\subsection{Social satisfaction}

Mei {\it et al.} prove that the \mymedian\ mechanism $\frac{3}{2}$-approximates the
optimal social satisfaction (Theorem 1 of \cite{flpdesire}).
In addition, they show that no deterministic and strategy proof
mechanism has an approximation ratio of the optimal social 
satisfaction of less than $8-4\sqrt{3}$ which is approximately 1.07
(Theorem 3 of  \cite{flpdesire}). 
\myOmit{
Indeed, the \mymidornearest\ mechanism 
provides such an $(8-4\sqrt{3})$-approximation of the optimal
social satisfaction with two agents. 
}
We cannot apply these results directly to our setting 
as the mechanisms considered in \cite{flpdesire} may select
an infeasible location for the facility. Instead, we again consider 
the \dmedian\ mechanism which locates the facility
at the nearest feasible location to the median agent. 

\begin{mytheorem} 
For a facility location problem with limited locations, 
the \dmedian\ mechanism is strategy proof and
has an unbounded approximation ratio of the optimal social satisfaction.
\end{mytheorem}
\myproof
Suppose the only feasible locations
for the facility are at 0 or 1, two agents are at $\frac{1}{2}$,
a third agent is at 1, and the tie-breaking
rule locates the facility at 0. This gives a social satisfaction of
zero units of happiness. However, the optimal social satisfaction of 1
unit of happiness requires
the facility to be located at $1$. We therefore have an unbounded approximation
ratio. Note that we can eliminate the dependency on the
tie-breaking rule by having the two agents not at $\frac{1}{2}$ 
but at $\frac{1}{2}-\epsilon$ 
for some small $\epsilon>0$ tending to zero. 
\myqed

Unfortunately, we cannot do better. 
No deterministic and strategy proof mechanism
has a bounded approximation ratio of the optimal social
satisfaction.

\begin{mytheorem} 
For a facility
location problem with limited locations, 
no deterministic and strategy proof mechanism has a bounded
approximation ratio of the optimal social satisfaction. 
\end{mytheorem}
\myproof
Suppose the only feasible locations
for the facility are at $0$ or $1$, and there
are two agents, one at $\frac{1}{2}-\epsilon$ and the other at
$\frac{1}{2}+\epsilon$
for small $\epsilon > 0$ that tends to zero. 
There are two cases. In the first case, 
the mechanism locates the facility at 0. 
Suppose the agent at $\frac{1}{2}+\epsilon$ mis-reports their
location as $1$. Since the mechanism is strategy
proof, the location of the facility cannot change.
Consider this new problem with one agent at 
$\frac{1}{2}-\epsilon$ and the other at $1$.
The solution returned by the mechanism 
has a social satisfaction that tends to zero units of happiness as
$\epsilon$ tends to zero. 
But the optimal social satisfaction is 1 unit of happiness when the 
facility is at $1$. 
Hence the mechanism has an unbounded approximation
ratio for the optimal social satisfaction. The second case where the facility
is located at $1$ is symmetric. 
\myqed

\subsection{Minimum satisfaction}

We turn now to the minimum satisfaction. 
\myOmit{
Mei {\it et al.} argue that the \mymidpoint\ mechanism $2$-approximates the
optimal minimum satisfaction (Section 3.3 of \cite{flpdesire}).}
Mei {\it et al.}  argue that no deterministic and strategy proof
mechanism has an approximation ratio of the optimal minimum
satisfaction of less than $\frac{4}{3}$ 
 (also Section 3.3 of  \cite{flpdesire}). 
The \mymedian\ mechanism has an unbounded approximation
ratio for the minimum satisfaction. Consider two agents at 0
and one at 1. Not surprisingly, the \dmedian\ mechanism may 
also approximate the optimal minimum
satisfaction poorly. 

\begin{mytheorem} 
For a facility
location problem with limited locations, the \dmedian\ mechanism
does not bound the approximation ratio of the optimal minimum satisfaction.
\end{mytheorem}
\myproof
Suppose facilities can be at 0, $\frac{1}{2}$ or 1. 
If there are two agents at 0 and one at 1 then
the \dmedian\ mechanism locates the facility at 0. 
The minimum satisfaction of this solution has zero units of happiness. However,
the optimal minimum satisfaction has $\frac{1}{2}$ unit of happiness
and is achieved by locating the facility at $\frac{1}{2}$. The
approximation ratio is therefore unbounded. 
\myqed

\myOmit{
We might have better hope for the performance of the \mydmidornearest\ mechanism.
When facilities are not limited in their location,
we can show that the \mymidornearest\ mechanism
$\frac{4}{3}$-approximates the 
optimal minimum satisfaction. 
This is optimal as no deterministic and strategy proof 
mechanism can do better \cite{flpdesire}. 
Unfortunately, the corresponding \mydmidornearest\ mechanism does not perform
as well when the facility is limited in where it can be located. 

\begin{mytheorem} 
For a facility
location problem with limited locations, the \mydmidornearest\ mechanism
does not bound the approximation ratio of the optimal minimum satisfaction.
\end{mytheorem}
\myproof
Suppose facilities can be at 0 or $\frac{5}{8}$. 
If there is one agent at 0 and a second at $\frac{3}{8}$ then
the \mydmidornearest\ mechanism locates the facility at $\frac{5}{8}$. 
The minimum satisfaction of this solution has zero units of happiness. However,
the optimal minimum satisfaction has 1 unit of happiness
and is achieved by locating the facility at 0. The
approximation ratio is therefore unbounded. 
\myqed
}

In fact, no deterministic and strategy proof mechanism
for a facility
location problem with limited locations has
a bounded approximation ratio of the minimum satisfaction.

\begin{mytheorem}
For a facility
location problem with limited locations, 
no deterministic and strategy proof mechanism has a bounded
approximation ratio of the optimal minimum satisfaction. 
\end{mytheorem}
\myproof
Suppose the approximation ratio is bounded by some factor $k$, 
and the feasible locations
for the facility are at $\frac{1}{2}-\frac{1}{k}$,
$\frac{1}{2}-\frac{1}{k}+\frac{1}{k^2}$,
$\frac{1}{2}+\frac{1}{k}-\frac{1}{k^2}$
or 
$\frac{1}{2}+\frac{1}{k}$, 
and there
are four agents, one at 
$\frac{1}{2}-\frac{2}{k}$,
$\frac{1}{2}-\frac{1}{k}+\frac{1}{k^2}$,
$\frac{1}{2}+\frac{1}{k}-\frac{1}{k^2}$
and the fourth at $\frac{1}{2}+\frac{2}{k}$.
Note that if the facility is located at 
$\frac{1}{2}-\frac{1}{k}$
or $\frac{1}{2}+\frac{1}{k}$
then the minimum satisfaction is zero
due to the agents at 
$\frac{1}{2}+\frac{2}{k}$ and
$\frac{1}{2}-\frac{2}{k}$ respectively. 
Hence, to have a bounded approximation ratio
the facility must be located 
at one of the optimal locations:
$\frac{1}{2}-\frac{1}{k}+\frac{1}{k^2}$ or
$\frac{1}{2}+\frac{1}{k}-\frac{1}{k^2}$.
There are two cases. 
In the first case, the facility is located at
$\frac{1}{2}-\frac{1}{k}+\frac{1}{k^2}$. 
Suppose the agent at 
$\frac{1}{2}+\frac{1}{k}-\frac{1}{k^2}$
changes their report of their
location to $1$. Since the mechanism is strategy
proof, the location of the facility must stay the same or more 
further away from the agent that moves. However, it cannot move further away as the
minimum satisfaction would drop to zero and the
approximation ratio would be unbounded. 
Consider this new problem with agents at 
$\frac{1}{2}-\frac{2}{k}$,
$\frac{1}{2}-\frac{1}{k}+\frac{1}{k^2}$,
$\frac{1}{2}+\frac{2}{k}$ and 1. 
As $k$ increases, the solution has a minimum satisfaction tending
to $\frac{2}{k^2}$ units of happiness. 
But the optimal minimum satisfaction tends to $\frac{4}{k}$ unit of happiness when the 
facility is at 
$\frac{1}{2}+\frac{1}{k}-\frac{1}{k^2}$. 
This gives an approximation ratio of $2k$, contradicting
the assumption that the approximation ratio is bounded by some
factor $k$. In the second case, the facility is located at
$\frac{1}{2}+\frac{1}{k}-\frac{1}{k^2}$. 
The argument is symmetric to the first case. 
\myqed

\section{Two facilities, distance approximations}

We move now from locating a single facility to locating two
facilities. When facilities are not limited in their location,
the only deterministic and strategy proof mechanism for locating two
facilities on the line
with a bounded approximation ratio for either the optimal total or 
maximum distance is the 
\myendpoint\ mechanism \cite{ft2013}. 
This provides a $(n-2)$-approximation of
the total distance and a 2-approximation of the
maximum distance. 

We suppose now that facilities are limited in their location, and 
consider the corresponding \mydendpoint\ mechanism that
locates the leftmost facility at the nearest feasible location
to the leftmost agent, tie-breaking to the right, 
and the rightmost facility at the nearest feasible location
to the rightmost agent, tie-breaking instead to the left. 
Tang {\it et al.}
prove that, when the two facilities are limited
in their locations, this mechanism $2n-3$-approximates the total distance \cite{flplimit2}.
Tang {\it et al.} also
prove that the mechanism $3$-approximates the maximum
distance, and that 
no deterministic and strategy proof mechanism can do better \cite{flplimit2}.

\section{Two facilities, welfare approximations}

When facilities are not limited in their location, 
the \myendpoint\ mechanism offers a good approximation of
both the optimal utilitarian and egalitarian welfare. 
In particular, it is not hard to show that it
$2$-approximates the optimal utilitarian welfare,
and $\frac{3}{2}$-approximates the optimal egalitarian welfare,

When facilities are limited in their locations, the corresponding
\mydendpoint\ mechanisms provides the same approximation
ratio for the utilitarian and egalitarian welfare.
We contrast this with the distance objective, where limiting the 
location of the two facilities worsens the approximation ratio
of the \mydendpoint\ mechanism.
We also contrast this with a single facility, where limiting the 
location of the facility worsens the approximation ratio
for either welfare objective. 

\begin{mytheorem}
The \mydendpoint\ mechanism 
$2$-approximates the optimal utilitarian welfare,
and $\frac{3}{2}$-approximates the optimal
egalitarian welfare.
\end{mytheorem}
\myproof
Suppose an agent has utility less than $\frac{1}{2}$. 
This is only possible iff the agent is served by a
facility at the nearest feasible location. 
Otherwise, an agent has an utility of $\frac{1}{2}$ or 
greater, compared to a maximum of 1. 
Hence, the total of their utilities, the utilitarian
welfare is at least half the optimal. 
Consider $n-2$ agents at $\frac{1}{2}$, and one each at 0 and 1. 
Suppose feasible locations are 0, $\frac{1}{2}$ and 1. Then 
the approximation ratio of the utilitarian welfare
is $\frac{2n-1}{n+2}$. As $n$ goes to
infinity, we can make this as close to 2 as we like. 

For the egalitarian welfare, there are two cases.
In the first case, suppose the agent traveling the furthest to be served
is served by the leftmost facility. There is a symmetric
argument in the other case for the rightmost facility. 
There are two subcases. In the first subcase, the leftmost facility is at or to
the right of the leftmost agent. The worst case for the
approximation ratio has agents at 0, 
$\frac{1}{2}$ and 1, 
with 0, $\frac{1}{4}$ and 1 being feasible locations for the facilities.
The \mydendpoint\ mechanism gives an approximation ratio of
$\frac{3}{2}$ of the optimal egalitarian welfare. 
In the second subcase, the leftmost facility is to the left of the
leftmost agent (as the leftmost agent is at an infeasible location
for the facility, and the nearest feasible location is to the left). 
The worst case for the approximation ratio has 
the agents at $\frac{1}{4}-\epsilon$, $\frac{1}{2}$ and 1,
with feasible locations for facilities at 0, $\frac{1}{2}$ and 1,
and $\epsilon$ approaching zero. 
The egalitarian welfare is maximized by locating facilities
at $\frac{1}{2}$ and 1.
The \mydendpoint\ mechanism gives therefore an approximation
ratio of $\frac{3}{2}$. 
\myqed

\section{Two facilities, satisfaction approximations}

When facilities are not limited in their location, 
the \myendpoint\ mechanism offers a good approximation of
the social satisfaction but not of the minimum satisfaction. 
In particular, it is not hard to show that it
$\frac{n}{2}-\frac{1}{4}$-approximates the optimal social satisfaction
but has no bound on the approximation ratio of 
the minimum satisfaction. 

When facilities are limited in their locations, the corresponding
\mydendpoint\ mechanism provides a larger approximation
ratio for the social satisfaction. 
We compare this with the distance objective, where limiting the 
location of the two facilities also worsens the approximation ratio
of the \mydendpoint\ mechanism.
We also compare this with a single facility, where limiting the 
location of the facility means that the approximation ratio
for the social satisfaction becomes unbounded. 

\begin{mytheorem}
The \mydendpoint\ mechanism 
$\frac{3n}{4}-\frac{1}{2}$-approximates the optimal social satisfaction, 
but does not bound the approximation ratio of the optimal
minimum satisfaction. 
\end{mytheorem}
\myproof
(Sketch) The worst case for the social satisfaction
has facilities at 0, $\frac{1}{2}$ or 1, $n-2$ agents
at $\frac{1}{2}$ and at $\frac{1}{4}-\epsilon$ and
$\frac{3}{4}+\epsilon$ for $\epsilon$ approaching zero.
The optimal facility location with respect to social 
satisfaction has a facility at $\frac{1}{2}$ and 0 or 1,
giving a social satisfaction tending to $n-\frac{2}{3}$. 
The \mydendpoint\ mechanism locates facilities at 0 and 1,
giving a social satisfaction tending to $\frac{4}{3}$. 
The approximation ratio is therefore $\frac{3n}{4} - \frac{1}{2}$. 

For the minimum satisfaction, consider agents at 0, $\frac{1}{2}$ and 1 with
facilities limited to these locations. 
The \mydendpoint\ mechanism locates
facilities at 0 and 1, giving a minimum satisfaction of 0.
This compares with a solution with optimal minimum
satisfaction that locates a facility at $\frac{1}{2}$, and
another at 0 or 1. The minimum satisfaction is now $\frac{1}{2}$
unit of happiness. Hence the approximation ratio is unbounded.
\myqed

\section{Conclusions}

We have studied the impact of constraints on the location of 
a facility on the performance of strategy proof mechanisms for
facility location. We considered six different objectives: 
the total and maximum distance agents must travel, 
the utilitarian and egalitarian welfare, and the social
and minimum satisfaction. In general, constraining facilities
to a limited set of locations makes all six objectives harder to
approximate in general. For example, a modified median mechanism  
is strategy proof and 3-approximates both the optimal
total and maximum distance. 
No deterministic and strategy proof mechanism can do better. This contrasts
with the setting in which there are no restrictions
on where facilities can be located and the median mechanism
returns the optimal total distance, and 2-approximates the optimal
maximum distance. 


\bibliographystyle{splncs}
\bibliography{/Users/tw/Documents/biblio/a-z,/Users/tw/Documents/biblio/a-z2,/Users/tw/Documents/biblio/pub,/Users/tw/Documents/biblio/pub2}

\end{document}
\end{document}